# scientific data

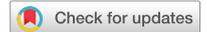

**OPEN**

**DATA DESCRIPTOR**

# N-Omniglot, a large-scale neuromorphic dataset for spatio-temporal sparse few-shot learning

Yang Li[1,2,6], Yiting Dong[1,3,6], Dongcheng Zhao[1,6] & Yi Zeng[1,2,3,4,5] ✉

Few-shot learning (learning with a few samples) is one of the most important cognitive abilities of the human brain. However, the current artificial intelligence systems meet difficulties in achieving this ability. Similar challenges also exist for biologically plausible spiking neural networks (SNNs). Datasets for traditional few-shot learning domains provide few amounts of temporal information. And the absence of neuromorphic datasets has hindered the development of few-shot learning for SNNs. Here, to the best of our knowledge, we provide the first neuromorphic dataset for few-shot learning using SNNs: N-Omniglot, based on the Dynamic Vision Sensor. It contains 1,623 categories of handwritten characters, with only 20 samples per class. N-Omniglot eliminates the need for a neuromorphic dataset for SNNs with high spareness and tremendous temporal coherence. Additionally, the dataset provides a powerful challenge and a suitable benchmark for developing SNNs algorithms in the few-shot learning domain due to the chronological information of strokes. We also provide the improved nearest neighbor, convolutional network, SiameseNet, and meta-learning algorithm in the spiking version for verification.

## Background & Summary

In recent years, large scale datasets and increased computing power have made machine learning, especially deep learning, reach a level of human-like performance in many areas[1–3]. However, compared with the human brain, artificial neural networks (ANNs) lack biological characteristic and interpretability, for their floating-point-based calculation and gradient-based algorithm[4]. Combining computer technology and computational neuroscience-related knowledge can effectively improve the current deep learning technology. Spiking neural networks (SNNs) are considered the third generation of artificial neural networks[5], by simulating similar calculations and representations in the human brain, which shows strong biological interpretability. Only neurons that fire spikes will participate in the calculation of the network. Meanwhile, the sparse spike activity greatly reduces the network's energy consumption[6]. However, the lack of datasets for SNNs burden the development of the SNN algorithm.

The success of deep learning can largely attribute to the introduction of datasets such as ImageNet[7] and COCO[8]. However, the currently widely used datasets are not suitable for SNNs. SNNs need to encode the static data into spike trains and then put them into the network[9]. As a result, the information will be missing, and it will not be fair to compare with the artificial neural networks. Dynamic Vision Sensor (DVS)[10] is a new neuromorphic camera. DVS only generates ON/OFF events on pixels with different light intensities to achieve low latency, low redundancy, and high time resolution, which is different from the frame-based cameras. In addition, DVS simulates the human visual nervous system in principle so that SNNs can fully use the temporal information provided by such sensors.

To better exploit the event properties of DVS, researchers have proposed many neuromorphic datasets using DVS. N-MNIST[11], N-caltech101 and DVS-CIFAR10[12] are obtained by using event cameras to record images from traditional classification datasets that follow predetermined or random trajectories of motion. In addition, researchers have tried to obtain events by recording activities in natural environments using neuromorphic cameras, such as DVS-Gesture[13] and N-Cars[14]. However, the existing datasets, such as those mentioned above, have very low temporal correlation, which indicates how much the data exhibits its characteristics over time. Low temporal correlation, i.e., high temporal redundancy, represents lower importance of the temporal dimension in judging the sample categories, and therefore does not facilitate our exploration of the spatio-temporal characterization capabilities of SNNs. All the characteristics are shown in Table 1. The average number of events per pixel at each time step is used to measure the sparsity, and the average cosine similarity between all pairs of frames for all samples is used to measure the difference. In addition to sparse coding[15] to reduce energy consumption, learning new concepts rapidly from a few samples is also one of the important capabilities of human brain. While it is an open problem in spike-based machine learning. The few-shot learning[16,17] imposes tremendous challenges on the current learning methodologies of SNNs due to the lack of neuromorphic datasets[18] for training and evaluating the learning ability of a few samples. Note that using artificial intelligence to process such high temporal resolution data for fast identification is still an open problem. However, it is crucial to use the event properties of neuromorphic data for biologically interpretable learning. Thus, we can provide a benchmark for improving the spatio-temporal information representation and few-shot learning capability of SNNs without encoding static data in a way that makes information lost.

[1]Brain-inspired Cognitive Intelligence Lab, Institute of Automation, Chinese Academy of Sciences, Beijing, China. [2]School of Artificial Intelligence, University of Chinese Academy of Sciences, Beijing, China. [3]School of Future Technology, University of Chinese Academy of Sciences, Beijing, China. [4]National Laboratory of Pattern Recognition, Institute of Automation, Chinese Academy of Sciences, Beijing, China. [5]Center for Excellence in Brain Science and Intelligence Technology, Chinese Academy of Sciences, Shanghai, China. [6]These authors contributed equally: Yang Li, Yiting Dong, Dongcheng Zhao. ✉e-mail: yi.zeng@ia.ac.cn

| Dataset | # of classes | Sparsity | Difference | Object |
|---|---|---|---|---|
| N-MNIST | 10 | 0.1466 | 0.3097 | Image |
| N-Caltech101 | 100 | 0.2118 | 0.3384 | Image |
| DVS-CIFAR10 | 10 | 0.4600 | 0.4561 | Image |
| N-Cars | 2 | 0.0107 | 0.3255 | Car |
| DVS-Gesture | 11 | 0.3356 | 0.7366 | Gesture |
| **N-Omniglot** | **1,623** | **0.0092** | **0.1256** | **Stroke** |

**Table 1.** The Characteristics of the neuromorphic datasets.

To tackle the problems and fulfill this gap, to the best of our knowledge, we propose the first neuromorphic dataset for few-shot learning using SNNs: N-Omniglot. The original Omniglot dataset[19] is the most commonly used dataset in the field of few-shot learning. It consists of 1,623 handwritten characters from 50 different languages. Each character has only 20 different samples. It is usually recognized as a static character image, while the rich temporal information of the writing process is ignored. Therefore, we reconstruct the writing process of strokes and use DVS to obtain the event records to get the neuromorphic version of Omniglot (N-Omniglot), as shown in Fig. 1. Various types of characters are expressed as event streams through the event camera, containing both temporal and spatial dimensions, thus providing a benchmark for spiking neural networks with binarized and rich spatio-temporal features. We provide several improved classic few-shot learning algorithms to adapt to SNN, showing that N-Omniglot varies in time dimension, provides more temporal information and supports many tasks. We hope it can provide a benchmark for SNN-based few-shot learning and provide a competitive environment for the research community to promote SNN's temporal and spatial feature extraction ability and sparse representation learning.

## Methods

In this work, we first use the stoke temporal information to reconstruct the writing process of Omniglot into videos. For the convenience of capturing, we merge the writing strokes of 20 samples of each character into a video file, and a blank sequence is inserted between each sample. Second, we use the DVS acquisition platform to shoot videos that are played on the monitor, and use the Robotic Process Automation (RPA) software to collect the data automatically. Finally, the corresponding sample data will be split out. Fig. 2 shows the entire construction process of the dataset.

**Stroke preprocessing and reconstruction.** Each image in Omniglot has corresponding stroke data in milliseconds. In order to present the entire writing process to the DVS, we first reconstruct the text record of strokes as a video of writing tracks. Also, because of the difference in acquisition equipment and writing habits, we delete the interval generated when each stroke is written. The linear interpolation algorithm is used to complete the data in milliseconds to reconstruct the character writing as accurately as possible. Due to the inconstant frequency of sampling and the jitter during the writing process, some strokes have only one or a few points, and the refresh rate of the display screen is 60 Hz, so strokes less than 17 ms in the reconstructed video may not be displayed. Therefore, we linearly interpolate them to a sufficiently long length. Here we use 34 ms.

**Automated capture using davis346.** We use Davis346 as our acquisition device due to its good time resolution. We design a black box to cover the screen and DVS camera to prevent external light changes from interfering with the experimental data collection. In the experiment, the DV software is used to process the captured event data. We set the background activity time parameter in the DV software to 4,000 to better filter out the background noise from the input, such as the low-frequency noise displayed on the monitor. As well as, the exposure parameter is fixed to 8000 to keep the brightness stable. In order to avoid frequent software operations that cause major changes in the relative position of the device, we use the Robotic Process Automation (RPA) software UiBot to automatically collect and record data. As shown in Fig. 2, we first read the address of the reconstructed stroke video. During the event conversion process, the record and stop buttons are pressed at the beginning and ending of the video playback. Finally, the recorded .aedat4 files are saved to the current directory.

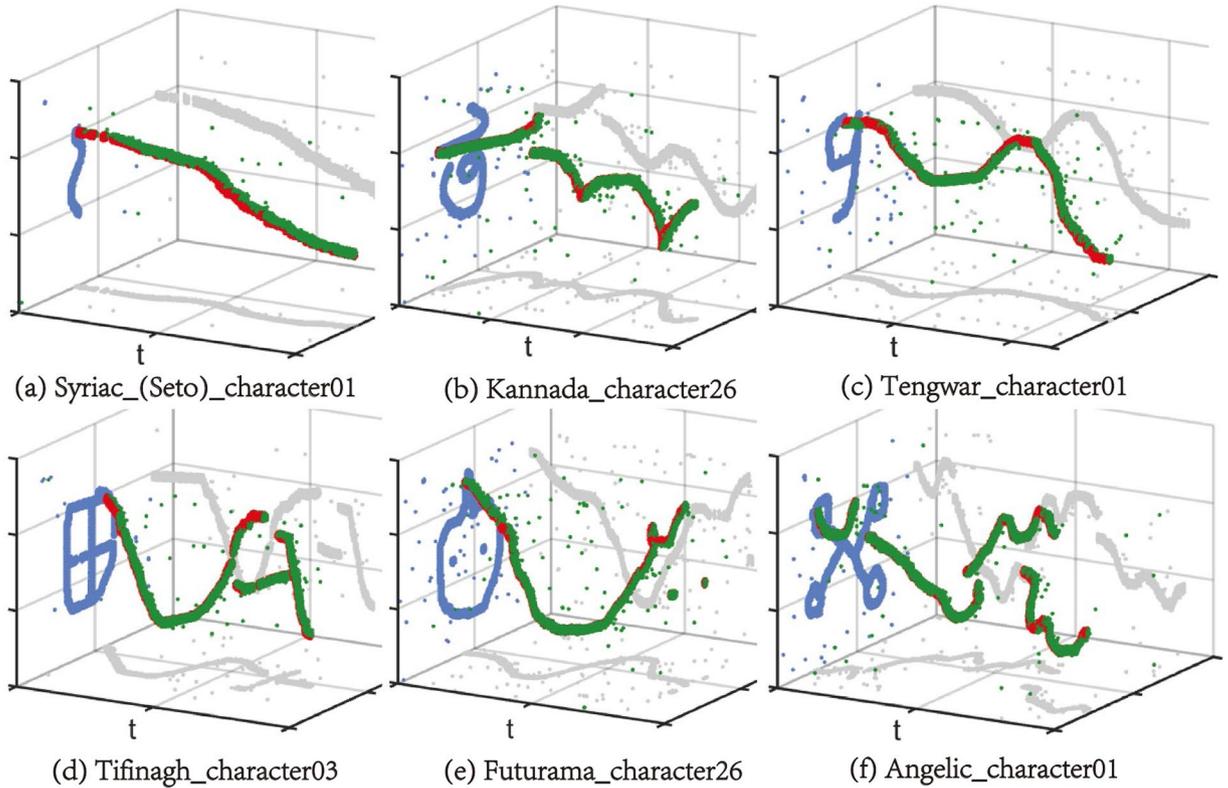

**Fig. 1** Some examples of N-Omniglot. High temporal correlation and spatio-temporal sparsity can be seen in these examples.

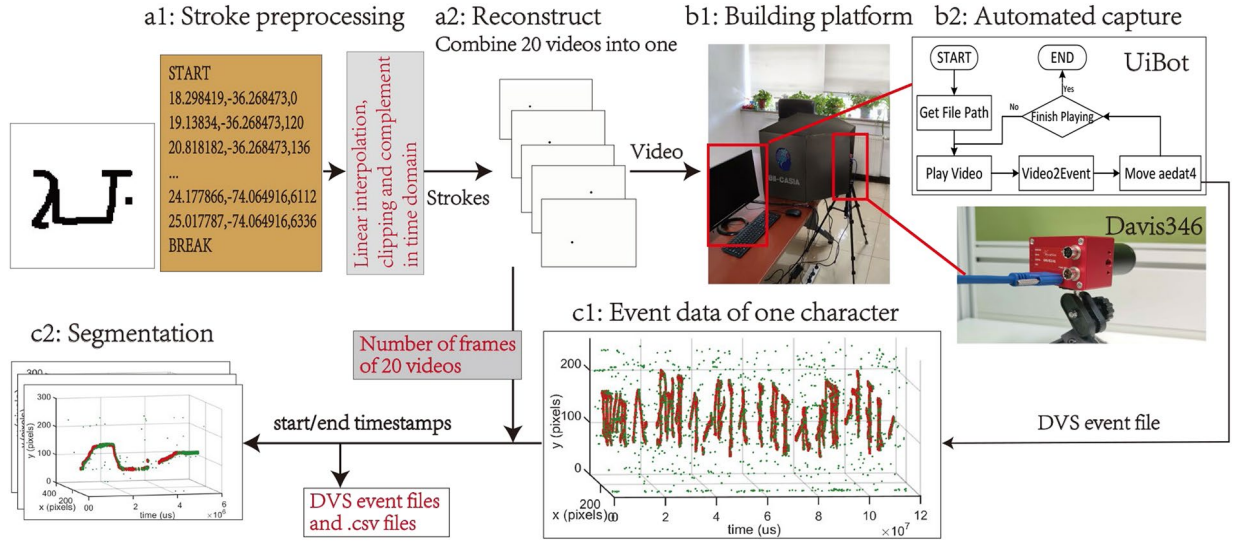

**Fig. 2** Complete process of data generation. Phase A constructs the video for recording, including **a1:** preprocessing the original data and **a2:** reconstructing the video. Phase B is the actual capture stage, including **b1:** building the equipment environment, and **b2:** recording with the RPA software. Phase C performs post-processing, including **c1:** labeling the beginning and end of the characters, and **c2:** segmentation using time stamp.

**Segmentation and preprocess for usage.** Each character in Omniglot contains 20 samples. To avoid unnecessary software operations and make the collected DVS event data more stable and efficient, we combine the 20 reconstructed videos into a long one, with a 500 ms gap between each sample, the beginning and the ending of the video. After converting the stroke video to the event file, we separate the events corresponding to each sample. Specifically, when the events change from sparse spatio-temporal property to concentrated in a certain position for a long time, the sample event starts. And we use the frame number of the original video to assist in finding the corresponding time node to determine the ending of the sample event. We save the event data with the form of (x, y, t, p), where the first two items x, y are the pixel coordinates of the event, the third item t is the timestamp of the event, and the fourth item p is the polarity with value 1 and 0 indicates the increase or decrease of brightness separately. In our paper, considering that SNN is always clock-driven at runtime and cannot perform asynchronous calculations on all neuron units like FPGAs, so the event data will be processed into image data to input into the network. We process the events within a period of time into an image with a resolution of DVS346 (346*260). The two polarities are represented by two channels, and the pixels without events are filled with 0. We mainly used two methods: the OR operation and the firing rates. The details will be shown in the technical validation section.

| Statistic | Mean | Std | Statistic | Mean | Std |
|---|---|---|---|---|---|
| X-max | 321.95 | 25.75 | ON events | 33,412.47 | 12,882.29 |
| X-min | 29.49 | 27.54 | OFF events | 31,606.19 | 12,111.33 |
| Y-max | 242.35 | 17.65 | #Language | 50 | — |
| Y-min | 8.55 | 16.03 | #Character | 1,623 | — |
| Time ($\mu s$) | 4,040,930 | 2,788,808 | #Sample | 32,460 | — |

**Table 2.** The statistical properties of the N-Omniglot dataset.

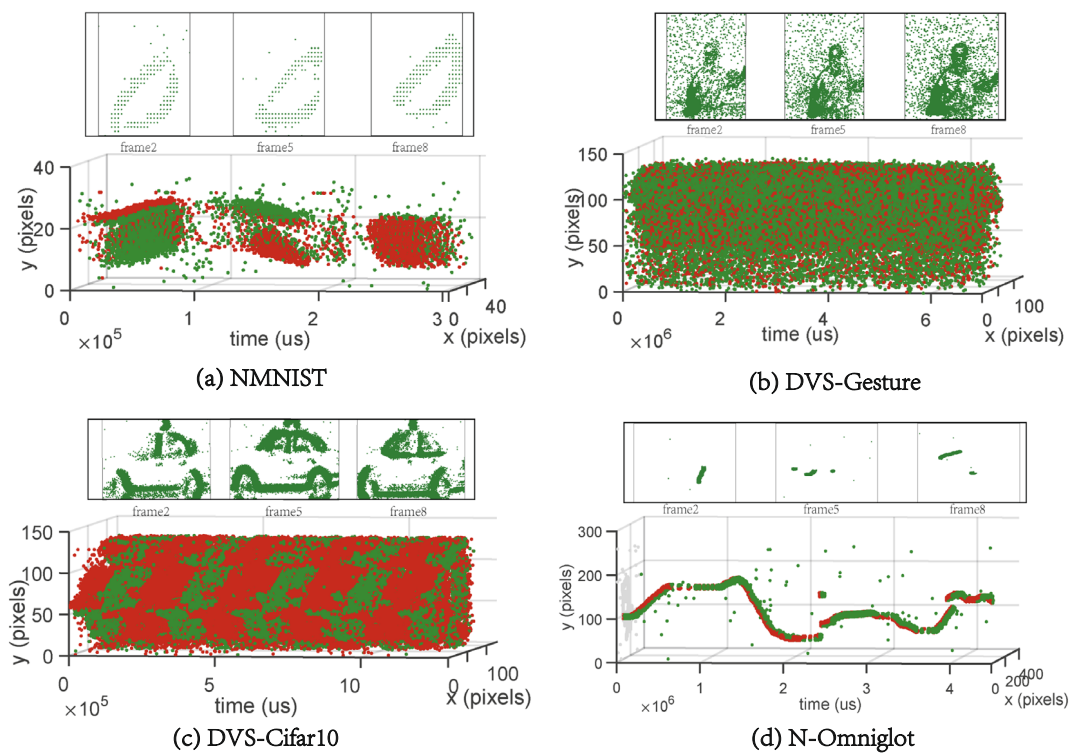

**Fig. 3** Comparison between N-Omniglot and other datasets. Four neuromorphic datasets are encoded into 12 frames, and frames 2, 5 and 8 are shown above.

**Visual analysis of N-Omniglot.** To better analyze the property of N-Omniglot, we calculate the statistical characteristics of N-Omniglot, as shown in Table 2. The maximum and minimum values of the horizontal and vertical coordinates are used to crop the image to remove unnecessary input. Also, the average writing time of the strokes of each sample is 4 s, which is realistic. In order to better illustrate the difference between the N-Omniglot dataset and other neuromorphic datasets, we visualize some samples in NMNIST, DVS-Gesture, DVS-CIFAR10 and N-Omniglot, as shown in Fig. 3. We use OR operation to compress all event data into 12 image frames, and select three frames of processed images. It can be seen that for the first three neuromorphic datasets, the three image frames are very similar, and the activity is very dense, whose difference is small compared with input the static images. However, the N-Omniglot dataset has large differences between frames and very sparse activities, which provides a greater challenge for building a high-performance spatio-temporal information processing algorithm, and the temporal information of the stroke order is crucial for the recognition and generation of characters. Significantly, this can also provide a good benchmark for the few-shot generation tasks.

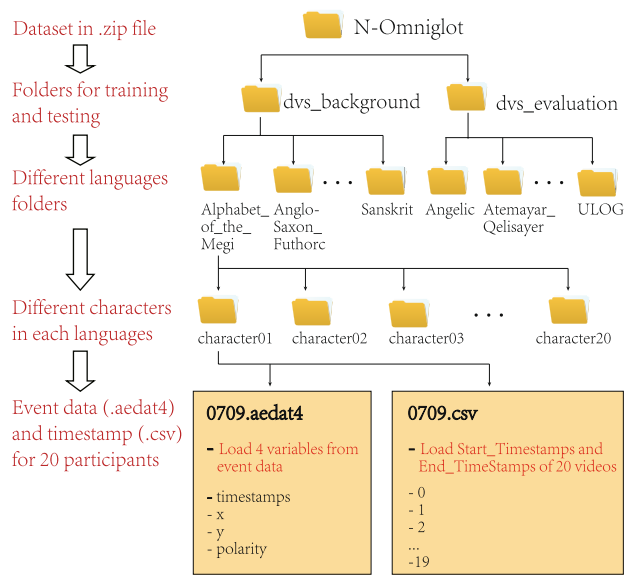

**Fig. 4** The structure of the N-Omniglot dataset.

## Data Records
The N-Omniglot dataset can be downloaded in Figshare[20] (https://doi.org/10.6084/m9.figshare.16821427). Because the data is collected using DV software, the stored file is aedat4 file. In addition, to minimize the additional operations of the acquisition process, we combine 20 samples into one video, using the method mentioned above to segment them. We save the beginning and ending timestamps of each sample in a CSV file. In order to better show the progress of our data compared with the previous data, we have displayed the event data on the homepage. We separate the files into different directories to facilitate maintenance and use while maintaining the same structure as Omniglot. As shown in the Fig. 4, the dataset contains 1,623 characters in 50 languages, and each character consists of 20 samples. The figure shows the retrieval process of the first character in the training set Alphabet_of_the_Megi. The aedat4 file records the entire event of the stroke instead of saving the color image frame. We also provide a preprocessed code at http://www.brain-cog.network/dataset/N-Omniglot/ to adapt it to the current algorithm application for easy use.

## Technical Validation
We notice that SNN is still in its infancy in the field of neuromorphic few-shot learning, and there is almost no suitable algorithm to support this task. In order to prove the effectiveness of N-Omniglot and the potential to provide new challenges for the training of SNN algorithms, experiments are conducted on four SNN algorithms, including two general classic pattern classification methods and two few-shot learning algorithms. Also, to demonstrate the difference between the N-Omniglot and the encoded Omniglot, all the experiments are performed on both. Then, we compare the differences between the encoded static dataset and the neuromorphic dataset.

**Encoding and preprocessing.** Static images do not contain temporal information. Therefore, to match the characteristics of SNN in processing spatio-temporal information, static images are usually processed into spike trains. As shown in Fig. 5a,c, we use poisson coding and constant coding[21,22] as the encoding strategy, which are the most commonly used in deep spiking neural networks. For the constant coding, the samples are fed directly into the network, and the first layer can be considered the encoding layer to generate the spike trains. On the other hand, for N-Omniglot, our dataset has temporal information and does not need to be encoded. However, neuromorphic datasets are acquired by DVS with the high temporal resolution, and the excessively long timeline is a huge burden for current clock-driven SNN algorithms. Therefore, datasets captured by DVS need to be merged in the temporal dimension. In our experiment, we divide the data by time average and combine them in two ways: the or operation and spike firing rate, as shown in Fig. 5b,d.

**Nearest neighbor.** As a classical pattern classification method, the nearest neighbor[23] (NN) method can evaluate the separability of samples to a certain extent and provide a benchmark for other algorithms to compare. As shown in Fig. 6c, the NN method compares the input sample with each in the training set and finds the sample closest to the input according to the given distance measurement function. Then the category of the input sample can be decided by the neighbor. In the experiment, we use the euclidean distance between different samples as the distance measure function.

**Classification directly.** The most considerable difficulty of few-shot learning lies in the large number of classification categories and a small number of samples per category. But even still, such tasks can be handled directly as general classification problems. Therefore, we construct a spiking convolutional neural network with leaky integrate-and fire (LIF) neurons to process the dataset. It consists of two convolution layers and two fully connected layers. Each convolutional layer is followed by an average pooling layer with step size 2. Due to the non-differentiable character of SNNs, we use the approximate gradient method of STBP[24] for training. Even if it is not specially designed for few-shot learning tasks, the algorithm can still obtain relatively efficient results. On the other hand, as a comparison, we also design an ANN network model, as shown in Fig. 5a, which uses the convolution layer as a feature extractor, while Long Short-Term Memory (LSTM)[25], which is a machine learning algorithm commonly used to extract time features, is employed to combine the information of temporal dimension.

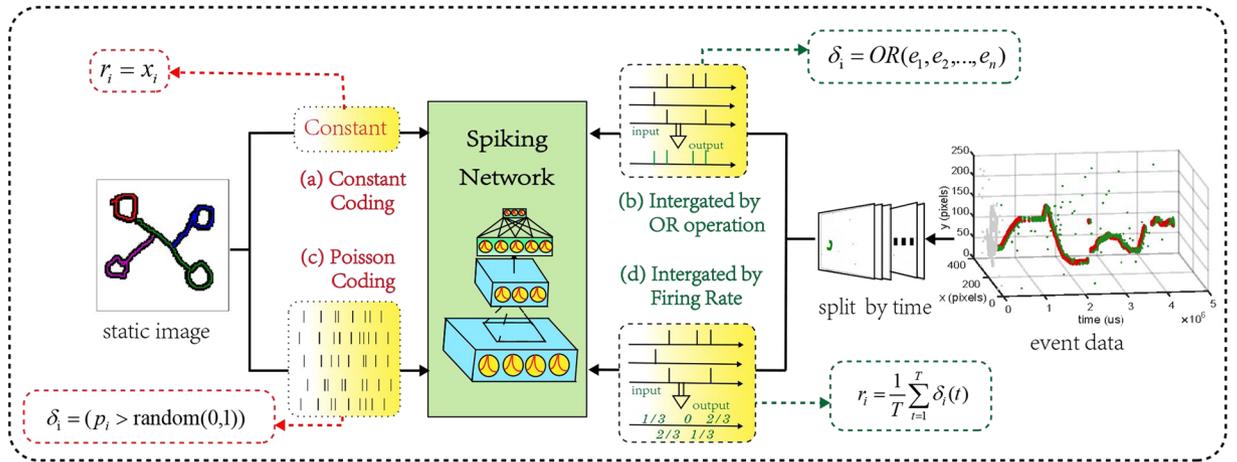

**Fig. 5** Encoding Method for Static Image and Preprocessing for Event Data. (**a**) Constant Coding and (**c**) Poisson Coding are used to encode static images, while (**b**) Event Frame and (**d**) Firing Rate are used to preprocess neuromorphic data.

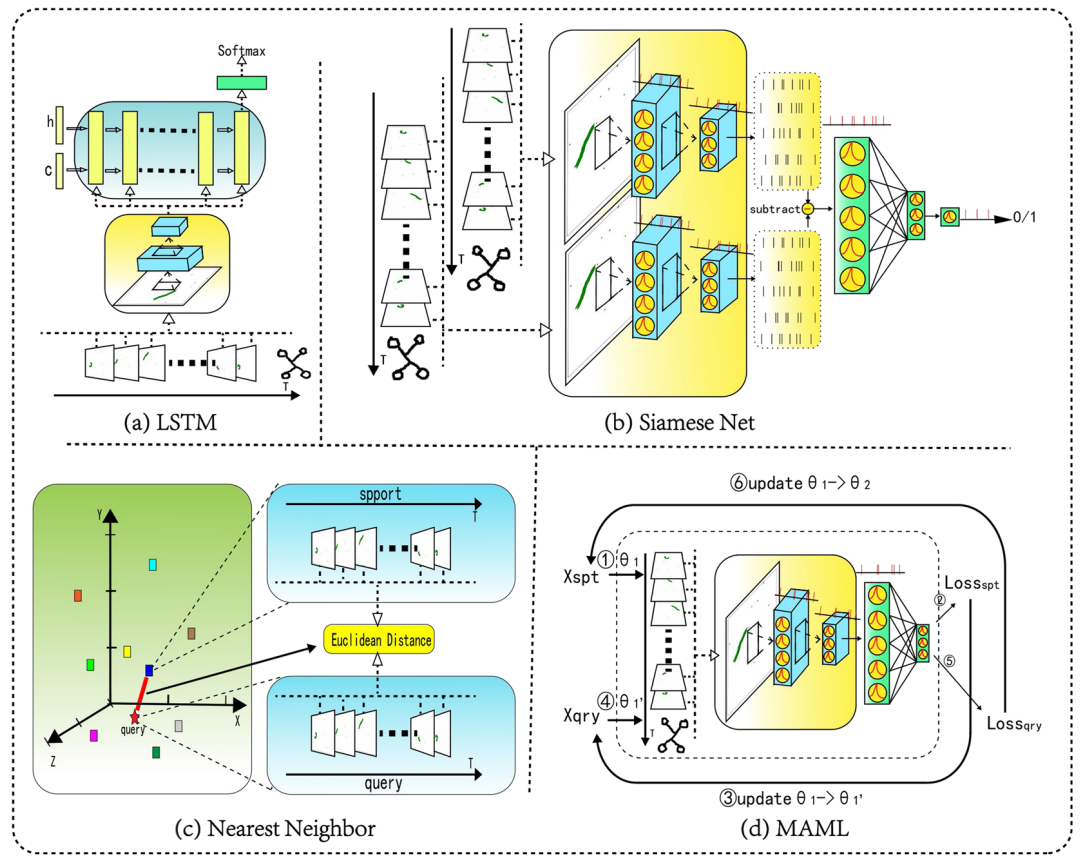

**Fig. 6** Four few-shot learning baseline methods for the proposed data descriptors. The image sequence represents the encoded frame. (**a**) The LSTM method is applied to ANN networks, and the corresponding SNN version replaces the LSTM layer with the fully connected layer. (**c**) The Nearest Neighbor method calculates the euclidean distance between samples as the classification basis. (**b**) The Siamese Net compares the spiking convolution feature representation of the two samples and uses the fully connected regression to classify their differences. (**d**) The MAML method optimizes the learning process from the gradient perspective to make the classifier have stronger generalization.

| Dataset | Methods | Structures | Encoding | Time | 20w1s | 20w5s | 5w1s | 5w5s |
|---|---|---|---|---|---|---|---|---|
| Omniglot | NN[23] | — | Constant | 12 | 17.6 ± 0.3% | 33.1 ± 0.5% | 37.5 ± 0.3% | 54.9 ± 0.4% |
| | | — | Poisson | 12 | 17.6 ± 0.6% | 32.3 ± 0.7% | 37.4 ± 0.5% | 55.1 ± 0.9% |
| | Direct | SCNN[1] | Constant | 12 | 63.2 ± 2.2% | | | |
| | | | Poisson | 12 | 64.6 ± 0.7% | | | |
| | | LSTM[2] | Constant | 12 | 66.3 ± 0.8% | | | |
| | | | Poisson | 12 | 67.0 ± 0.8% | | | |
| | Siamese Net[26] | SCNN[3] | Constant | 12 | 74.9 ± 0.5% | 90.4 ± 0.8% | 89.1 ± 0.5% | 97.0 ± 0.3% |
| | | | Poisson | 12 | 75.3 ± 0.4% | 90.0 ± 0.4% | 89.3 ± 0.4% | 97.1 ± 0.1% |
| | MAML[27] | SCNN[4] | Constant | 12 | 80.7 ± 0.3% | 94.8 ± 0.5% | 91.5 ± 0.1% | 97.4 ± 0.3% |
| | | | Poisson | 12 | 83.3 ± 0.7% | 94.8 ± 0.2% | 93.0 ± 0.2% | 97.8 ± 0.1% |
| N-Omniglot | NN[23] | — | Event | 12 | 7.4 ± 0.4% | 10.6 ± 0.2% | 26.7 ± 0.2% | 32.2 ± 0.5% |
| | | — | Frequency | 12 | 7.3 ± 0.3% | 10.6 ± 0.3% | 23.3 ± 0.4% | 27.6 ± 0.3% |
| | Direct | SCNN[1] | Event | 12 | 62.6 ± 0.8% | | | |
| | | | Frequency | 12 | 60.8 ± 0.9% | | | |
| | | LSTM[2] | Event | 12 | 46.2 ± 0.3% | | | |
| | | | Frequency | 12 | 43.3 ± 0.3% | | | |
| | Siamese Net[26] | SCNN[3] | Event | 4 | 54.0 ± 0.7% | 76.5 ± 0.6% | 73.1 ± 0.6% | 89.0 ± 0.4% |
| | | | Event | 8 | 51.4 ± 0.8% | 73.6 ± 0.3% | 70.4 ± 0.5% | 87.2 ± 0.4% |
| | | | Event | 12 | 50.8 ± 1.2% | 73.0 ± 1.2% | 69.9 ± 0.9% | 87.0 ± 1.0% |
| | | | Frequency | 4 | 53.3 ± 1.6% | 74.9 ± 1.6% | 72.6 ± 0.9% | 88.4 ± 1.0% |
| | | | Frequency | 8 | 50.8 ± 0.4% | 72.2 ± 0.8% | 69.8 ± 0.3% | 86.6 ± 0.7% |
| | | | Frequency | 12 | 49.8 ± 1.3% | 71.3 ± 1.0% | 69.3 ± 0.8% | 85.7 ± 0.6% |
| | MAML[27] | SCNN[4] | Event | 4 | — | — | 77.2 ± 0.4% | 91.5 ± 0.5% |
| | | | Event | 8 | — | — | 72.8 ± 0.7% | 89.3 ± 0.7% |
| | | | Event | 12 | — | — | 70.7 ± 0.4% | 88.9 ± 0.6% |
| | | | Frequency | 4 | — | — | 74.4 ± 0.7% | 90.7 ± 0.1% |
| | | | Frequency | 8 | — | — | 71.1 ± 0.3% | 88.6 ± 0.4% |
| | | | Frequency | 12 | — | — | 70.3 ± 0.9% | 87.3 ± 0.3% |

**Table 3.** Accuracy of various modified classical methods on N-Omniglot. [1]SCNN: 15c5-AP2-40c5-AP2-300-1623. [2]ANN: (32c3-MP2)*2-LSTM128-1623. [3]SCNN: 64c10-128c7-4096-1. [4]SCNN: 15c5-AP2-40c5-AP2-300-5/20.

**Siamese net.** The Siamese Net[26] is a classical few-shot learning algorithm based on metrics. Because the original Siamese Net could not handle the neuromorphic dataset, we improve it by using SNN with LIF neurons as the basic network to add the ability to process temporal information for the model. The Siamese Net inputs two samples at the same time. If the sample pairs belong to the same category, they are marked as 1; otherwise, they are marked as 0. The network compares the two samples to determine whether they belong to the same category. As shown in Fig. 6b, the two samples share the first half of the network structurally, and the difference between the two feature maps is input into the later fully connected layer. During the test phase, the given query set is compared with the samples of the support set one by one, and the category with the largest probability value is output as the classification result.

**MAML.** MAML[27] (Model-Agnostic Meta-Learning) is another classic few-shot learning algorithm based on optimization. For the same reason, we improve MAML into SNN version, exploiting the SNN with LIF neurons as the network backbone to make the model capable of processing neuromorphic datasets. MAML tries to gain the ability to converge after a few iterations quickly. First, a fixed number of classes are randomly drawn from the dataset, with a fixed number of samples from each category, including support and query sets. Then the weights are copied and updated several times over the support set in the training phase. The loss calculation is performed with the copied weights on the query set, and the original network weight is updated with the corresponding gradient. While the same operation is performed on the support set as in the training phase and directly outputs the classification results on the query set in the test phase.

**Experimental result.** Table 3 shows the classification accuracy results of Omniglot and N-Omniglot under the four methods. Different preprocessing methods and simulation time lengths are used to compare the experimental results. The last four columns of the table represent four typical configurations for few-shot learning. N-way K-shot indicates that the support set sampled from the dataset consists of N classes, and each class consists of K samples during the testing phase. Similar to Omniglot, the result of NN and direct classification methods on N-Omniglot is much better than random guesses, proving the validity of the proposed data descriptor. As shown in the table, the performance of the four methods on N-Omniglot is lower than the results on Omniglot. The first reason is that the proposed dataset is more sparse in the spatial dimension. The similarity of data in the temporal dimension is lower than the input based on static image or poisson coding, which brings a new challenge to SNN learning. Another reason is the lack of preprocessing methods for neuromorphic datasets. It can be seen from the table that the preprocess method based on the event frame has better performance than that based on firing rate. When using firing rate, due to data sparseness in the spatial and temporal dimensions, the difference of floating-point values synthesized by preprocessing is significant and not conducive to spikes' generation and transmission. Therefore, new requirements are needed for the SNN preprocessing methods. It is worth noting that we simultaneously test the identification accuracy of the two classical few-shot learning methods at different simulation times. The results show that the longer the simulation time, the lower the accuracy. It is because the longer the simulation time, the more frames the event is divided into, and the more difficult it is to connect information between frames. It indicates that the data descriptor is essential for improving SNN's ability to extract more important spatio-temporal features. Therefore, the N-Omniglot proposed in this paper can be considered an effective, robust, and challenging dataset.

## Usage Notes

We provide three data interfaces for N-Omniglot to meet the requirements of different algorithms for data loading. In addition, four improved SNN learning methods, used in the technical verification section for N-Omniglot, can be found on http://www.brain-cog.network/dataset/N-Omniglot/. The researcher needs to download the dataset[20] from https://doi.org/10.6084/m9.figshare.16821427 and merge the folders into two (dvs_background and dvs_evaluation). The program will first read the aedat4 file and split the file contents when first using the code. Then a folder is created with the same structure as Omniglot, containing all the NumPy format samples, so that no more time will be spent here later. According to the different preprocessing methods, data with different frames can be obtained and directly input into the neural network. Researchers can directly use these four improved SNN few-shot learning algorithms to process datasets, or develop new algorithms to preprocess N-Omniglot datasets and propose novel few-shot learning algorithms suitable for neuromorphic datasets.

## Code availability

Preprocessing code for the dataset and few-shot learning algorithms to verify its quality can be found here: https://github.com/Brain-Cog-Lab/N-Omniglot. The code uses Python3 and PyTorch platforms, and the Torchvision package version is expected to be higher than 0.8.1. Please refer to the Usage Notes section and ReadMe file to run the code.

### Acknowledgements

This study is supported by National Key Research and Development Program (Grant No. 2020AAA0104305), the Strategic Priority Research Program of the Chinese Academy of Sciences (Grant No. XDB32070100).

### Author contributions

L.Y., Z.D. and Z.Y. came up with the idea, L.Y. and D.Y. performed the experiment and analysed the results. L.Y., D.Y., Z.D. and Z.Y. wrote the paper.

### Competing interests

The authors declare no competing interests.

### Additional information

**Correspondence** and requests for materials should be addressed to Y.Z.

**Reprints and permissions information** is available at www.nature.com/reprints.

**Publisher's note** Springer Nature remains neutral with regard to jurisdictional claims in published maps and institutional affiliations.

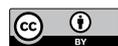